\pdfoutput=1

\documentclass[11pt]{article}

\usepackage[preprint]{acl}

\usepackage{epsfig}
\usepackage{amsmath}
\usepackage{amssymb}
\usepackage[utf8]{inputenc} 
\usepackage[T1]{fontenc}    
\usepackage{url}            
\usepackage{booktabs}       
\usepackage{amsfonts}       
\usepackage{nicefrac}       
\usepackage{microtype}      
\usepackage[misc]{ifsym}
\usepackage{multirow}
\usepackage{mathtools}
\usepackage{amsthm}
\usepackage{comment}
\usepackage{subfigure}
\usepackage{enumerate}
\usepackage{threeparttable}
\usepackage{enumitem}
\usepackage{graphicx}
\usepackage{times}
\usepackage{latexsym}
\usepackage{subcaption}
\usepackage{verbatimbox}
\usepackage{longtable} 
\usepackage{bbm}
\usepackage{colortbl} 
\usepackage{pifont} 
\title{Argus: Benchmarking and Enhancing Vision-Language Models for 3D Radiology Report Generation}

\author{
Che Liu\textsuperscript{1}, 
Zhongwei Wan\textsuperscript{2}, 
Yuqi Wang\textsuperscript{3},
Hui Shen\textsuperscript{2}, \\
\bf Haozhe Wang\textsuperscript{4},
\bf Kangyu Zheng\textsuperscript{2}, 
\bf Mi Zhang\textsuperscript{2}, 
\bf Rossella Arcucci\textsuperscript{1} \\
\textsuperscript{1}Imperial College London, 
\textsuperscript{2}The Ohio State University, 
\textsuperscript{3}University of Hong Kong, \\
\textsuperscript{4}Hong Kong University of Science and Technology \\
\href{mailto:che.liu21@imperial.ac.uk}{che.liu21@imperial.ac.uk} \\
}

\begin{document}
\maketitle
\begin{abstract}
Automatic radiology report generation holds significant potential to streamline the labor-intensive process of report writing by radiologists, particularly for 3D radiographs such as CT scans. While CT scans are critical for clinical diagnostics, they remain less explored compared to 2D radiographs. To date, there has been no comprehensive benchmark for 3D radiograph report generation (3DRRG), nor sufficient investigation into the optimal training strategies for Vision Language Models (VLMs) in this context, particularly with respect to vision encoder choices, visual token compression, and model scaling.
In this work, we make three key contributions. We curate \textbf{CT-3DRRG}, the largest \textbf{publicly} available 3D CT-report dataset, establishing a robust and diverse benchmark for evaluating VLM performance on 3DRRG. Furthermore, we propose a comprehensive training recipe for building high-performing VLMs for 3DRRG, exploring key factors such as vision encoder pretraining strategies, visual token compression, and the impact of data \& model scale. Guided by these findings, we introduce \textbf{Argus}, a state-of-the-art family of VLMs that achieve superior performance across different model sizes and input 3D medical image resolutions, efficiently processing high-resolution 3D images up to $512 \times 512 \times 256$\footnote{All code and data will be released after acceptance.}.

\end{abstract}

\section{Introduction}


Radiology reports are essential for clinical decision making, yet their manual creation is labor-intensive and time-consuming \cite{bastawrous2017improving, rimmer2017radiologist, rosenkrantz2016us}. This has driven the need for automation in radiology report generation. Existing work on this task primarily focuses on 2D images, such as chest X-rays \cite{tanida2023interactive, li2023unify}. However, compared to 2D images, 3D medical images provide more comprehensive diagnostic information and are crucial for identifying life-threatening diseases such as pulmonary opacities and early-stage cancer \cite{bradley2019sensitivity, self2013high}.

While 3D radiograph report generation (3DRRG) has the potential to significantly improve clinical workflows, several challenges remain unresolved in the literature. Notably, there is no comprehensive benchmark for 3DRRG, and many existing studies lack robust evaluations across multiple datasets and metrics. Traditional NLP metrics alone are inadequate for assessing the clinical relevance of generated reports. As a result, new evaluation metrics, such as clinical efficacy measures like GREEN~\cite{ostmeier2024green} and RaTEScore~\cite{zhao2024ratescore}, are essential for inclusion in the benchmark to evaluate report quality in clinical settings.

Additionally, high-resolution (HR) 3D images offer more clinical value than low-resolution (LR) images, as certain conditions, such as pulmonary nodules, are harder to detect in LR scans~\cite{liu2021application}. However, most studies~\cite{bai2024m3d,wu2023towards} downsample HR volumes (e.g., $512 \times 512 \times 256$) to LR ($256 \times 256 \times 32$) for report generation, which leads to significant information loss and may overlook critical details. This downsampling is primarily driven by the increased computational cost associated with processing the larger number of visual tokens generated by HR volumes.
Furthermore, no clear training recipe exists for effectively building Vision-Language Models (VLMs) for 3DRRG, particularly in terms of selecting and pretraining vision encoders, visual token compression, and optimizing model and data scale.

This brings us to a crucial challenge: \textbf{the lack of large-scale public datasets and a comprehensive benchmark} for 3DRRG. Additionally, \textbf{training strategies for VLMs remain unclear}, requiring a systematic analysis of key components.

\begin{figure*}[t!]
    \centering
\includegraphics[width=0.99\textwidth]{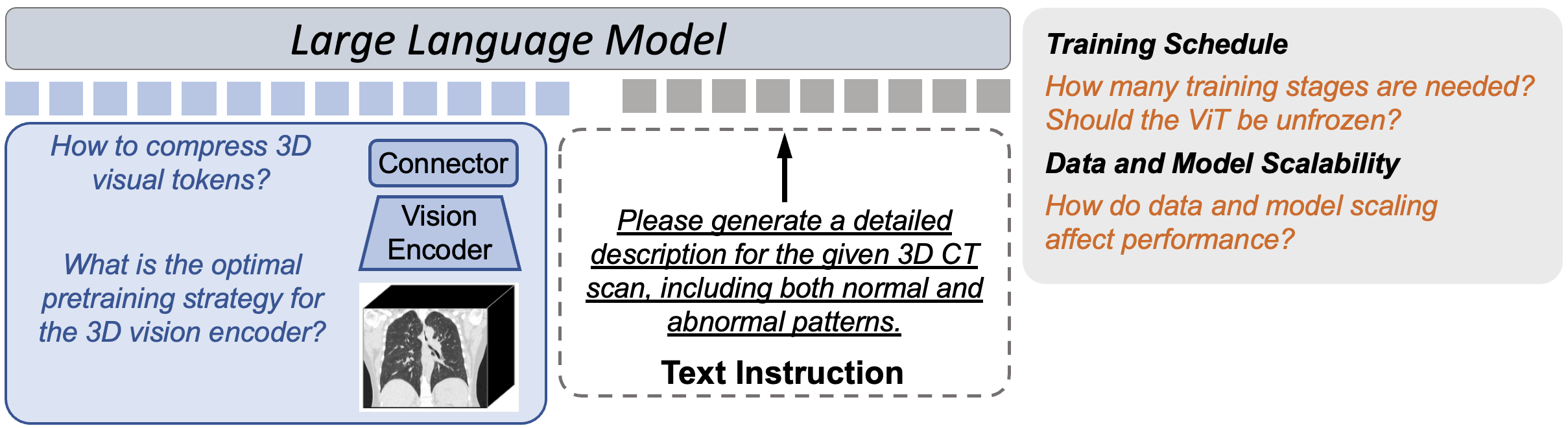}
\vspace{-10pt}
    \caption{\textbf{Argus framework}. A schematic illustration of our comprehensive exploration of 3DRRG, covering key design choices from 3D vision encoder pretraining and visual token compression to training schedules and scalability. We systematically analyze each component in Sections \ref{sec:vlm design} and \ref{sec:train schedule}.}
    \label{fig:framework}
\vspace{-20pt}
\end{figure*}

To address this challenge, we make 4 key contributions:
\begin{itemize}
    \item We curate the largest publicly available 3D CT-report dataset, \textbf{CT-3DRRG}, comprising 49,849 CT volumes paired with radiology reports from three publicly available sources: BIMCV-R, CT-RATE, and INSPECT~\cite{chen2024bimcv,ct-rate,huang2023inspect}.

    \item Based on CT-3DRRG, we establish the first comprehensive benchmark for 3DRRG, incorporating 8 NLP metrics and 3 clinical efficacy metrics. This enables broad evaluation across diverse sources, moving beyond single-resource or in-house data.

    \item We conduct an in-depth analysis of the key components required to build VLMs for 3DRRG, including strategies for \textbf{vision encoder pretraining} and \textbf{visual token compression}, the \textbf{VLM training schedule}, and the impact of \textbf{data and model scaling}.

    \item We propose a family of VLMs, \textbf{Argus}, that achieves superior performance on 3DRRG, surpassing existing methods in both NLP and clinical efficacy metrics. Argus scales from 3B to 70B parameters and efficiently handles 3D image resolutions up to $512 \times 512 \times 256$. 

\end{itemize}

\section{Related Works}

\noindent\textbf{Vision-Language Models.} 
The development of Vision-Language Models (VLMs) combines computer vision and natural language processing to enhance visual and linguistic capabilities. This integration is crucial for tasks that require both visual perception and language comprehension. Models like CLIP \cite{radford2019language}, Flamingo \cite{alayrac2022flamingo}, and BLIP \cite{li2023blip} have improved the alignment between these modalities by using extensive image-text data samples, resulting in significant performance gains.
LLaVA \cite{liu2023visual} further simplifies this approach by using a basic linear projector to align visual features with language space, involving only a few learnable parts and tailored instruction data to fully leverage the model's strong capabilities. Despite their effectiveness, these approaches focus mainly on 2D images.
To explore the relatively underexplored area of 3D medical imaging, such as CT scans, we extend the LLaVA framework by incorporating a customized 3D vision encoder specifically designed for 3D image variants. Our goal is to tackle the challenge of generating accurate radiology reports for 3D medical images.

\noindent\textbf{Radiology Report Generation for 2D Medical Images.}
Radiology report generation for 2D images typically involves template-based generation with region differentiation~\cite{li2018hybrid, jing2017automatic}, knowledge integration and coherence to highlight abnormalities~\cite{liu2021exploring, ma2021contrastive}, and cross-modal alignment using attention to link textual and visual features~\cite{gu2024complex,tanida2023interactive, li2023unify}.
Although effective for 2D settings (e.g., chest X-rays), extending these methods to 3D volumetric data (e.g., CT scans) is challenging. The 3D context involves distinct patterns in each slice (e.g., lungs and heart in upper slices versus abdominal organs in lower slices), whereas 2D images typically present uniform spatial structures with intensity variations~\cite{adegun2021deep, puttagunta2021medical, singh20203d}.
\\
\begin{figure*}[ht]
    \centering
    \includegraphics[width=0.99\textwidth]{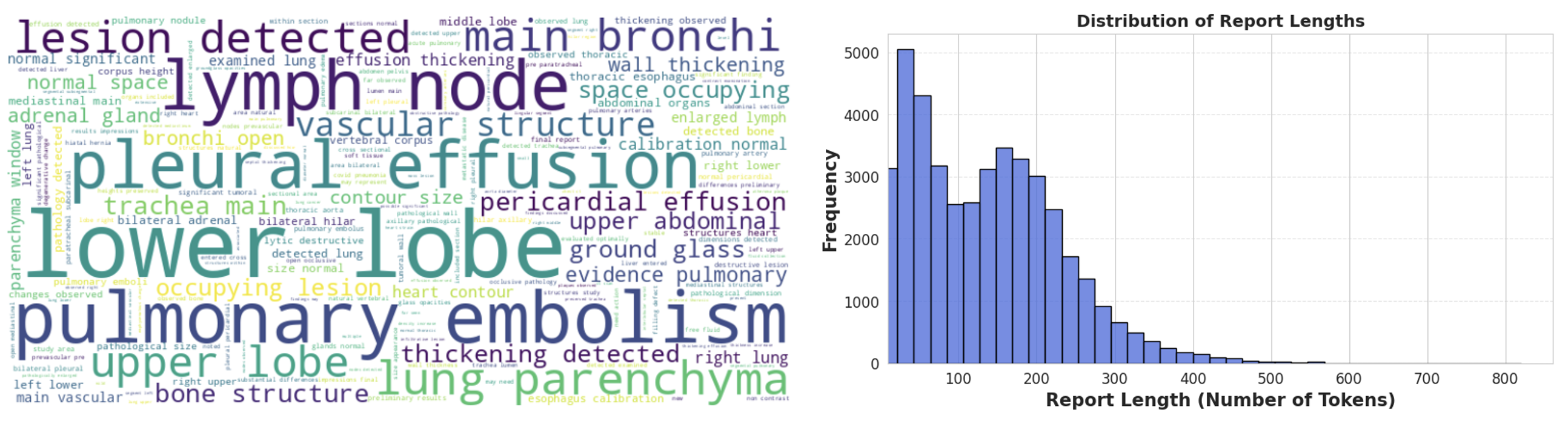}
    \caption{Distribution of the CT-3DRRG dataset. \textbf{Left}: Word cloud visualization highlighting the most frequent terms in the radiology reports. \textbf{Right}: Histogram showing the distribution of report lengths (measured in number of tokens).}
    \label{fig:data dist}
\end{figure*}
\noindent\textbf{Radiology Report Generation for 3D Medical Images.} 
Despite its clinical significance, only a few studies~\cite{hamamci2024ct2rep, wu2023towards, bai2024m3d} have attempted 3D radiograph report generation (3DRRG) with publicly available code, while others~\cite{lai2024e3d, chen2024large} do not release their code, hindering direct comparisons. Current methods have notable drawbacks: CT2Rep~\cite{hamamci2024ct2rep} processes high-resolution (HR) volumes slice by slice, losing 3D context; RadFM~\cite{wu2023towards} and M3D~\cite{bai2024m3d} downsample HR inputs to low-resolution (LR) views and use Preceiver~\cite{alayrac2022flamingo} to compress large numbers of visual tokens to just 64 or apply spatial pooling, potentially corrupting critical visual information. Preserving HR is crucial because downsampling coronal/sagittal axes can obscure subtle abnormalities~\cite{bradley2019sensitivity, self2013high}, and clinical practice often favors 256-slice CT scans for enhanced diagnostic detail~\cite{klass2010prospectively, chua2013diagnostic}. Moreover, these approaches rely on small-scale or permission-restricted datasets, limiting their broader evaluation and generalizability.

To advance 3DRRG, we curate the largest publicly accessible 3D CT-report dataset for evaluating VLM performance. We also conduct a thorough investigation of crucial components in VLM training, enabling our approach to process 3D scans at up to $512 \times 512 \times 256$ resolution without downsampling, thus preserving essential diagnostic details.

\section{Benchmark Construction}
\label{sec:dataset}

\subsection{Curating the CT-3DRRG Dataset}
To establish a robust and diverse benchmark for 3D Radiology Report Generation (3DRRG), we construct the CT-3DRRG dataset by integrating three publicly available resources: BIMCV-R~\cite{chen2024bimcv}, CT-RATE~\cite{ct-rate}, and INSPECT~\cite{huang2023inspect}. To ensure data quality, we apply a systematic curation process to the original datasets, as described in Section ~\ref{sec: curate}.

Notably, we retain the entire CT-RATE official test set without filtering. After curation, the number of samples in each dataset is summarized in Table~\ref{tab:split}. The BIMCV-R and INSPECT datasets are split into training, validation, and test sets following a 70\%/10\%/20\% split. For the CT-RATE dataset, we utilize its official test set and allocate 10\% of the training set as a validation set. Thus, the CT-3DRRG dataset consists of three subsets corresponding to these sources.

Following curation, we propose CT-3DRRG, a unified dataset constructed by merging multiple publicly available datasets. The statistical details of CT-3DRRG are presented in Table~\ref{tab:split} and Figure~\ref{fig:data dist}. During training, we combine the training sets from all sources into a single dataset. For evaluation, we evaluates VLM performance on the test set of each subset separately, allowing for a thorough analysis of robustness and generalizability across different data distributions.

\begin{table}[t!]
  \centering
  \scalebox{0.7}{
  \begin{tabular}{l|c|c|c}
    \toprule[1.2pt]
    \textbf{Dataset}  & \multicolumn{1}{c|}{\textbf{Train/Val/Test}} & \multicolumn{1}{c|}{\textbf{Avg. Tokens }} & \multicolumn{1}{c}{\textbf{Max Tokens}} \\
    & \textbf{Split} & \textbf{per Report} & \textbf{per Report} \\ 
    \midrule
    BIMCV-R           & 3,726/532/1,064                     & 97.19                               & 406                              \\
    CT-RATE           & 21,715/2,412/1,564                     & 201.40                               & 824                              \\
    INSPECT           & 14,280/2,040/4,080                     & 48.71                               & 267                              \\
    \hline
    \textbf{Total}    & 39,721/4,984/5,144                     & 130.03                               & 824                              \\
    \bottomrule[1.2pt]
  \end{tabular}
  }
    \caption{Processed data details of CT-3DRRG.}
  \label{tab:split}
\end{table}

\subsection{Data Preprocessing}
After building the CT-3DRRG dataset, we preprocess the data for training and evaluating the VLM. The preprocessing steps are designed to standardize the data and ensure it is in the appropriate format for model training.

\textbf{HU Value Clipping and Intensity Normalization:} The Hounsfield Unit (HU) values of the CT scans are clipped to the range of -1000 to +1000 to standardize the intensity levels. Then, the intensity values are normalized to the [0, 1] range to ensure uniform scaling across all images.

\textbf{Spatial Normalization:} Following the RadGenome-Chest CT~\cite{zhang2024radgenome}, the spacing of the CT scans is normalized to [1, 1, 4] mm on the sagittal, coronal, and transverse axes to ensure consistency in voxel spacing across samples.

\textbf{Resizing:} All CT scans are resized to two different resolution settings. For high-resolution settings, the scans are resized to $512 \times 512 \times 256$, where 512 represents the dimensions of the sagittal and coronal axes, and 256 is the number of slices in the transverse axis. For normal-resolution settings, the scans are resized to $256 \times 256 \times 64$, where 256 refers to the size in the sagittal and coronal axes, and 64 is the number of slices in the transverse axis.

To preserve the integrity of spatial relationships, which is crucial for accurate interpretation of 3D medical images, augmentation techniques such as flipping or rotating were not employed. This decision was made to avoid potential ambiguities in spatial orientation, particularly given radiological reports that specify locations such as the ``left lung" and ``right lung."

For the instructions used during training the VLM, we utilize the instruction: \textit{``Please generate a detailed description for the given 3D CT scan, including both normal and abnormal patterns."}. Inspired by \cite{bai2024m3d} where diverse instructions are employed to prevent overfitting on a specific instruction style. To further enhance diversity, we use GPT-4o to rephrase the instruction into 30 distinct variants, all maintaining the same semantic meaning.

\subsection{Selecting Appropriate Metrics}
For the 3DRRG task, the most straightforward evaluation approach involves using NLP-based metrics such as BLEU, METEOR, CIDEr, and ROUGE~\cite{bai2024m3d}. However, these metrics primarily assess lexical and syntactic similarity while overlooking the clinical correctness and relevance of the generated reports. Some studies \cite{ct-rate,ct2rep} attempt to address this limitation by employing text classifiers, such as those in CT2Rep~\cite{ct2rep}, to evaluate report accuracy. However, these classifiers are restricted to limited predefined categories, making them unsuitable for more diverse datasets like BIMCV-R~\cite{chen2024bimcv} and INSPECT~\cite{huang2023inspect}. To provide a clinically meaningful assessment, we adopt RadGraph-XL~\cite{radgraphxl}, GREEN~\cite{ostmeier2024green}, and RaTEScore~\cite{zhao2024ratescore} as evaluation metrics. These metrics offer a more reliable evaluation by focusing on structured clinical information, factual consistency, and radiological correctness, leveraging specifically trained models for clinical assessment.

\section{Exploring VLM Design for 3DRRG}
\label{sec:vlm design}
\subsection{Vision Encoder Pre-training}
\label{sec: vision pretrain}
\begin{figure*}[ht!]
    \centering
    \includegraphics[width=0.99\textwidth]{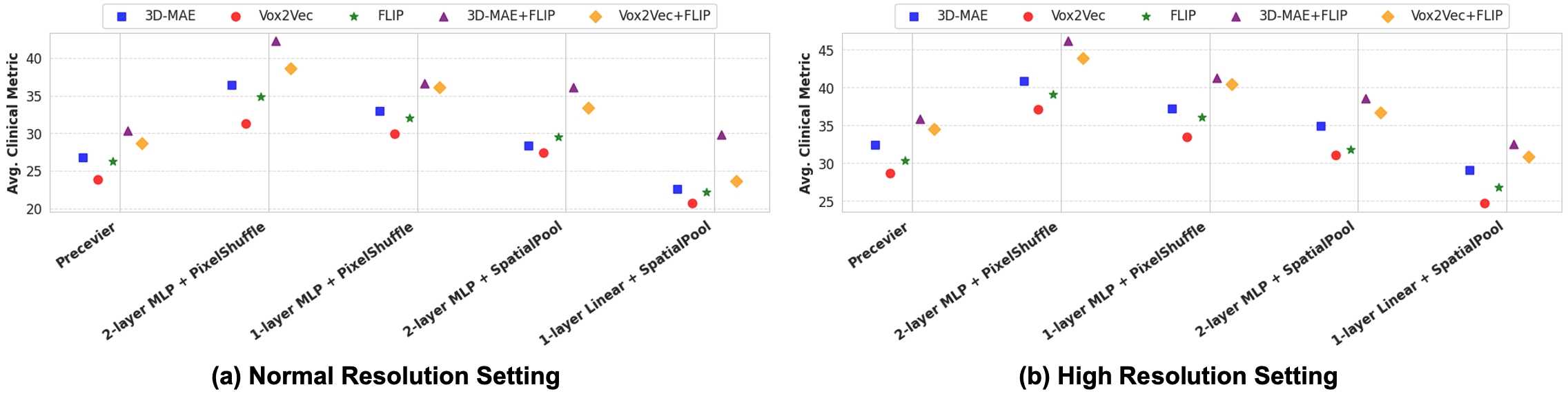}
    \caption{Comparison of different model combinations involving various vision encoder pretraining strategies, visual token compression, and connector designs under \textbf{(a)} normal resolution and \textbf{(b)} high resolution settings. The results represent the average clinical metric across the three subsets mentioned in Section \ref{sec:dataset}.}
    \label{fig:visualpretrain}
\end{figure*}

In this section, we investigate the pre-training strategy of the vision encoder for the VLM. We use the ViT-Base~\cite{dosovitskiy2020image} architecture, incorporating 3D positional embeddings and patching to adapt it for 3D CT volumes. To assess the impact of resolution, we explore two settings: for normal-resolution CT scans ($256\times256\times64$), we use a patch size of $16\times16\times8$, generating 2048 visual tokens; for high-resolution scans ($512\times512\times256$), a patch size of $32\times32\times16$ produces 4096 tokens. These configurations allow us to analyze how resolution influences the encoder’s ability to learn meaningful visual representations. 
To comprehensively evaluate the impact of different pre-training strategies on the vision encoder, we explore a range of visual self-supervised learning (vSSL) methods and language supervision techniques.

\begin{figure}[t]
    \centering
    \includegraphics[width=0.9\linewidth]{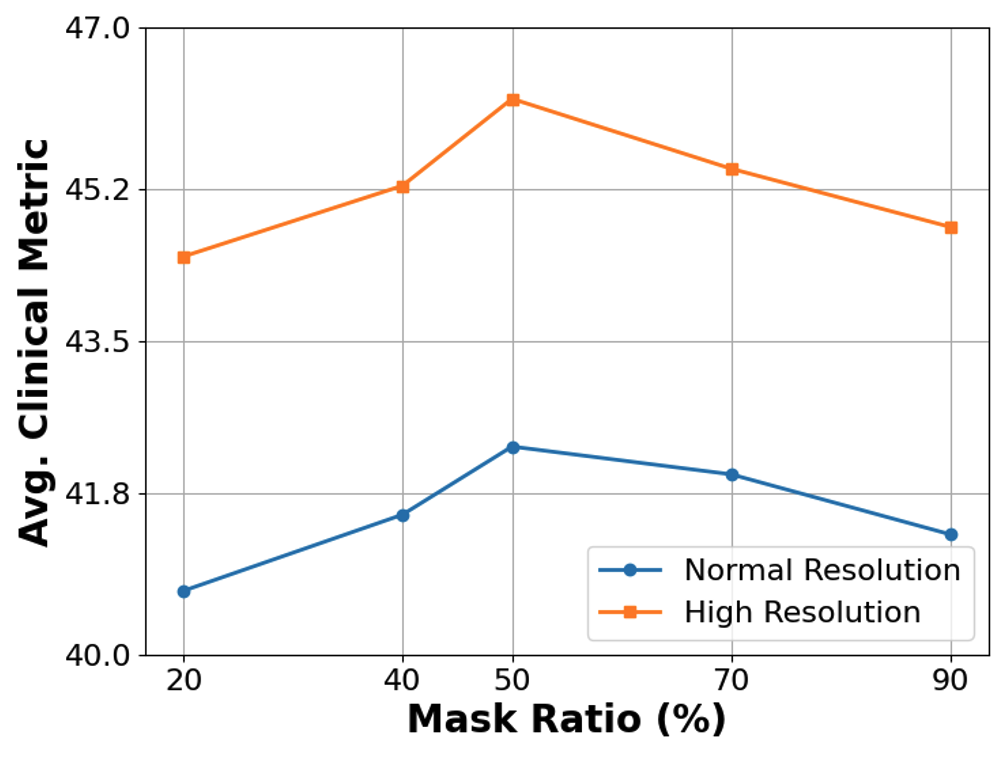}
    \caption{Ablation study on the masking ratio for vision encoder pretraining. We evaluate the impact of the mask ratio with 3D-MAE combined with the FLIP strategy using a 8B LLM.}
    \label{fig:maskratio}
    \vspace{-10pt}
\end{figure}

\begin{itemize}
    \setlength{\itemsep}{0pt}
    \item \textbf{3D-MAE \cite{3dmae}}: Masks visual tokens to reconstruct voxel intensities, learning visual representations.
    \item \textbf{Vox2Vec \cite{vox2vec}}: Aims to learn consistent voxel-level representations across views, enhancing voxel-wise representation consistency.
    \item \textbf{FLIP \cite{flip}}: Pretrains the visual encoder with patch masking, aligning visual representations with textual descriptions.
    \item \textbf{Vox2Vec + FLIP}: Trains in two stages: (1) learns voxel-wise representations with Vox2Vec, (2) applies contrastive learning to align 3D CT with reports.
    \item \textbf{3D-MAE + FLIP}: Trains in two stages: (1) reconstructs masked patches with 3D-MAE, (2) applies contrastive learning to align 3D CT with reports.
\end{itemize}
For methods involving patch masking, we set the masking ratio to 50\%. Additionally, we analyze different masking ratios in Figure \ref{fig:maskratio} to assess their impact on model performance.

\subsection{Visual Token Compression and Connector}
\label{sec: token compress}
In this section, we explore various strategies for visual token compression and the corresponding connectors, as the large number of visual tokens generated from 3D CT scans increases the cost of visual token processing in the VLM. In M3D~\cite{bai2024m3d}, a 3D average pooling is applied to the ViT output visual tokens, and MLP is used as a connector to convert the visual token dimensions to the LLM embedding dimensions. Meanwhile, RadFM~\cite{wu2023towards} uses a Preceiver to compress these tokens into 64 randomly initialized learnable queries. However, these methods all result in some loss of visual information. To address this, we introduce pixel shuffle~\cite{shi2016real,team2024internvl2} as a non-information-loss compression method to investigate the optimal combination of visual token compression and connectors.
For both pooling and pixel shuffle, we set the downsampling factor to 0.5 on each dimension. This means that for normal resolution, 2048 tokens are downsampled to 256 tokens, and for high resolution, 4096 tokens are downsampled to 512 tokens.

As shown in Figure \ref{fig:visualpretrain}, we combine these token compression strategies on the features extracted from the vision encoder, pretrained with various methods mentioned in Section \ref{sec: vision pretrain}. The results show that the 2-layer MLP with pixel shuffle, using a vision encoder pretrained with 3D-MAE + FLIP, outperforms the other methods across all metrics. This provides the following key findings:
\begin{figure}[t]
    \centering
    \includegraphics[width=0.99\linewidth]{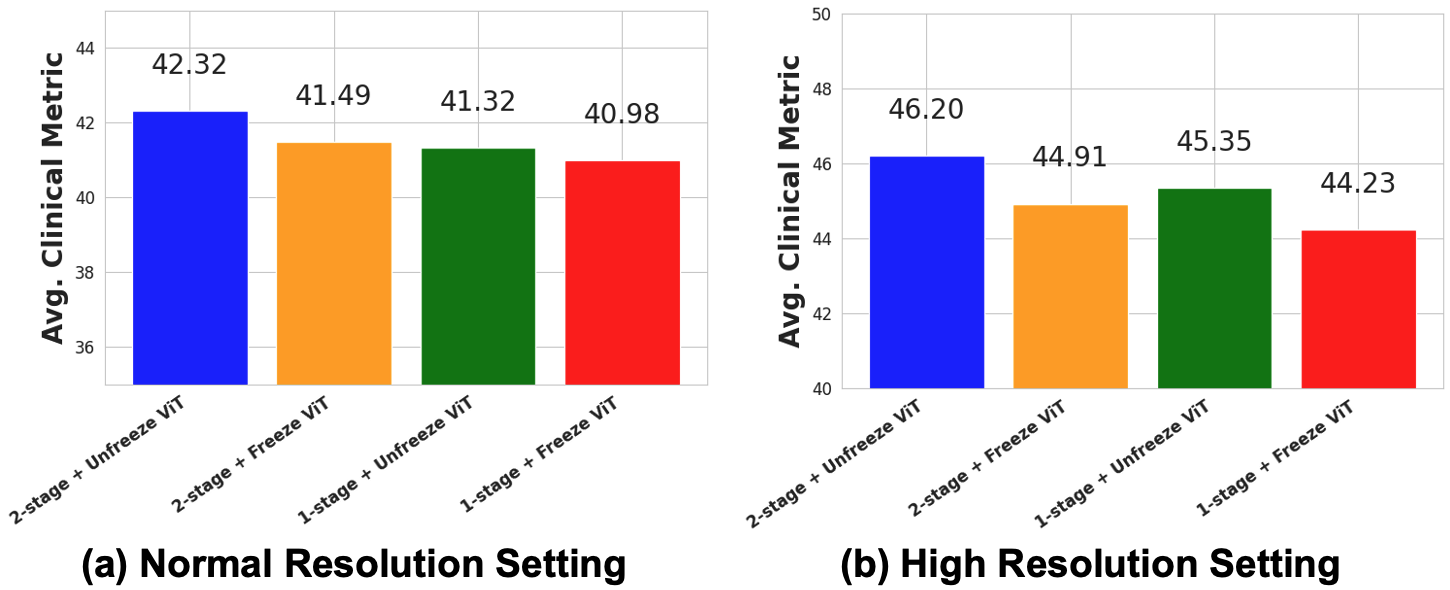}
    \caption{Comparison of different training schedules, including multi-stage strategies and variations of ViT, under \textbf{(a)} normal resolution and \textbf{(b)} high resolution settings. The results reflect the average clinical metric computed across the three subsets outlined in Section \ref{sec:dataset}.}
    \label{fig:trainschedule}
\end{figure}
\begin{enumerate}
    \item \textbf{3D-MAE combined with language supervision achieves superior performance.} This is because 3D-MAE does not rely on data augmentation techniques, unlike Vox2Vec~\cite{vox2vec}, which uses rotation or flipping that can distort the spatial relationships in 3D CT scans. Additionally, the two-stage visual pretraining helps progressively enhance the model’s ability to learn more accurate and meaningful representations.

    \item \textbf{Pixel shuffle is more effective than other compression methods, as it does not result in information loss.} Pixel shuffle reduces the number of tokens while increasing the dimensionality of each token, retaining all visual information. In contrast, pooling performs brute-force averaging of all pixels, which can lead to a loss of spatial detail. The Preceiver method, using random query initialization to compress all visual tokens into only 64 queries, results in substantial information loss, as discussed in \cite{yao2024deco}. Additionally, the 2-layer MLP with various compression methods consistently outperforms the 1-layer MLP.
\end{enumerate}

Based on these insights, we recommend that the optimal VLM should train the vision encoder using visual SSL combined with language supervision, utilize pixel shuffle for visual token compression, and adopt a 2-layer MLP as the connector strategy.

\section{How to Properly Train a VLM for 3DRRG?}
\label{sec:train schedule}
\subsection{Training Schedules}

For our VLM training, we select Llama3.1-instruct as the LLM backbone for the 8B and 70B models, and Llama3.2-instruct for the 3B models. For the vision encoder and token compression strategies, we follow the findings presented in Sections \ref{sec: vision pretrain} and \ref{sec: token compress}. 

In this section, we explore the impact of different training schedules. In the \textbf{1-stage} approach, we jointly train the connector and the LLM for one epoch using supervised fine-tuning (SFT). The \textbf{2-stage} approach consists of two steps: first, we freeze the ViT and LLM while training only the connector for one epoch; second, we train the connector, MLP, and LLM together using SFT for one epoch. We use the DeepSpeed ZeRO-3 offload setting for all model scales to ensure consistency in our experiments. The hyperparameters are detailed in Section \ref{sec: hyper training}

As depicted in Figure \ref{fig:trainschedule}, we find that the 2-stage approach consistently outperforms the 1-stage method, aligning with the findings in \cite{tong2024cambrian}. Additionally, we investigate the effect of freezing versus unfreezing the ViT. Our experiments indicate that unfreezing the ViT consistently enhances performance, particularly under the high-resolution setting, where it achieves the highest performance. This suggests that allowing the ViT to update its weights refines feature representations, leading to improved 3DRRG performance. Given these findings, we recommend using the 2-stage training approach with an unfrozen ViT to further boost performance.

\subsection{Investigating the Effects of Data and Model Scales}
In this section, we explore how scaling both data and model size affects VLM training performance. As shown in Figure \ref{fig:scale}, we begin with a dataset of 10K samples and progressively increase the size to 20K, 30K, and finally the full dataset. The results demonstrate a consistent improvement in performance as the dataset size increases, highlighting the model's scalability. We evaluate this effect on both normal-resolution and high-resolution settings.

Additionally, we examine the impact of scaling the LLM size from 3B to 70B parameters, as shown in Figure \ref{fig:scale}. The results indicate that larger models yield better performance, further validating the scalability of our training strategy. These findings suggest that our approach is effective and can seamlessly scale to accommodate both larger datasets and more powerful LLMs.

\section{Argus, the State-of-the-Art VLM for 3DRRG}

\begin{figure*}[ht!]
    \centering
    \begin{minipage}{0.99\textwidth}
        \centering
        \includegraphics[width=\textwidth]{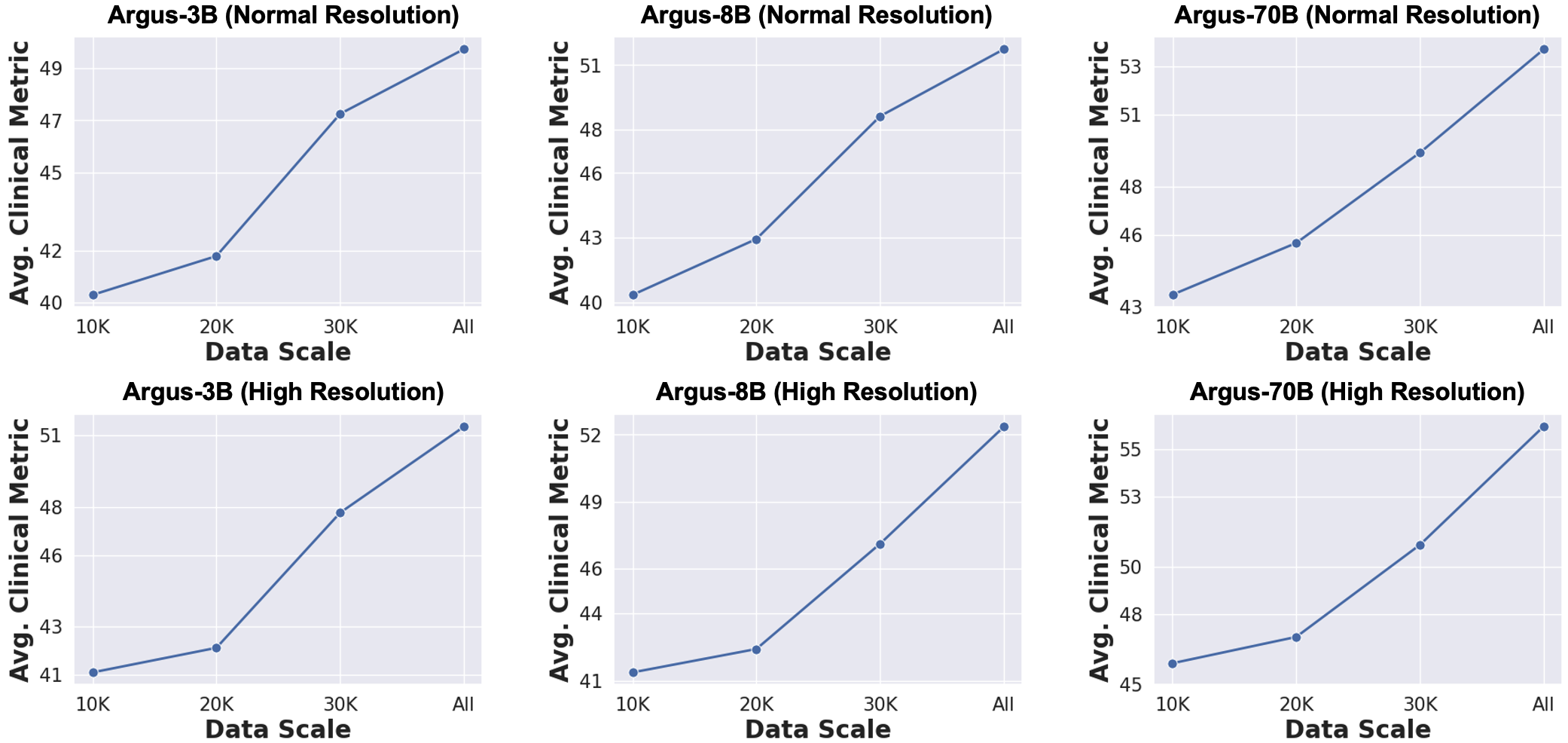}
        \caption{Scaling study of Argus with different LLM sizes and data scales. We evaluate Argus at 3B, 8B, and 70B parameters under both normal and high-resolution settings while increasing the data scale from 10K to the full dataset. The results show a consistent improvement in performance as both model size and data scale increase, demonstrating the robustness and scalability of our approach.}
        \label{fig:scale}
    \end{minipage}
    
    \vspace{0.5cm} 

    \begin{minipage}{0.99\textwidth}
        \centering
        \includegraphics[width=\textwidth]{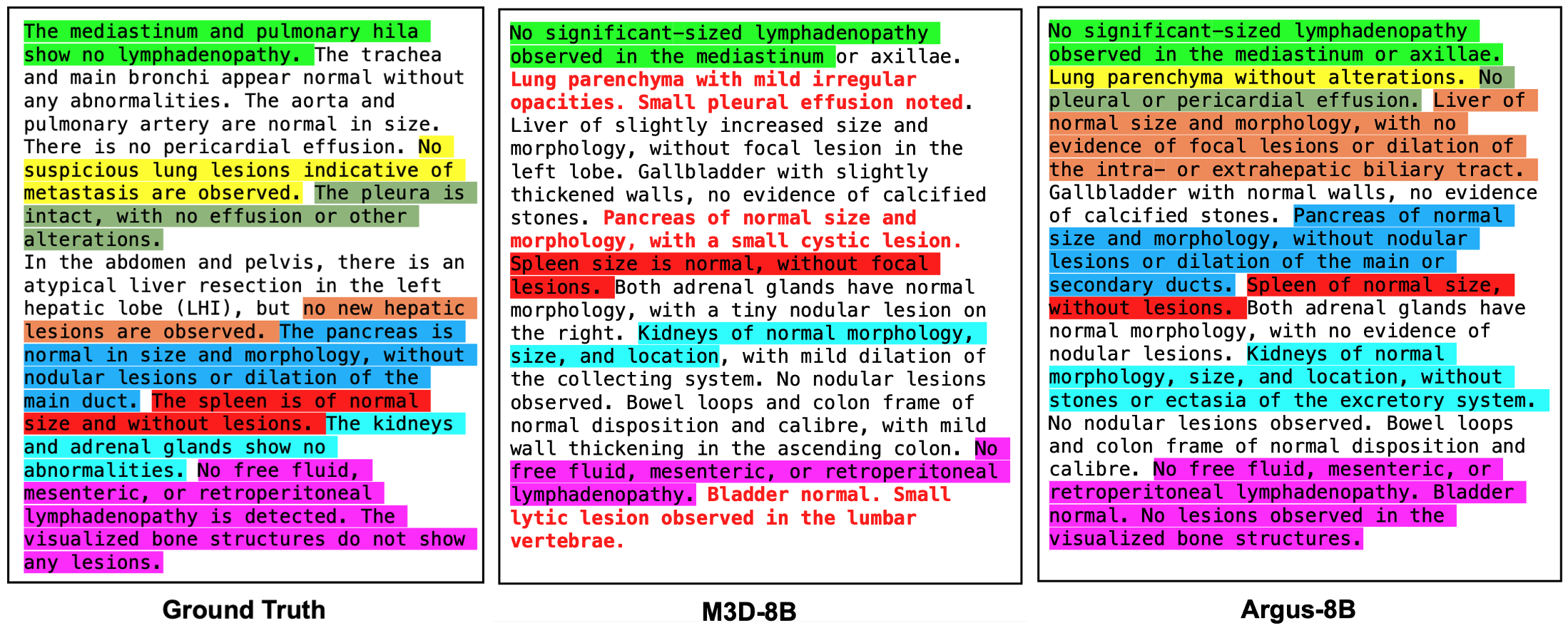}
        \caption{Qualitative comparison between the ground truth and reports generated by existing methods (M3D-8B) and our method (Argus-8B). Highlighted background text indicates correctly generated content, while bold red text denotes incorrect information. 
        M3D-8B exhibits inconsistencies in lesion description and misidentifies key anatomical structures, whereas Argus-8B generates more accurate and clinically relevant descriptions, demonstrating its superiority in medical report generation.}
        \label{fig:report}
    \end{minipage}
    \vspace{-20pt}
\end{figure*}

\begin{table*}[h]
  \centering
  \scalebox{0.85}{
  \begin{tabular}{l|cccc|cccc|cccc}
    \toprule[1.2pt]
    \textbf{Model} 
    & \multicolumn{4}{c}{\textbf{CT-RATE}} 
    & \multicolumn{4}{c}{\textbf{BIMCV-R}} 
    & \multicolumn{4}{c}{\textbf{INSPECT}} \\
    & {\fontsize{7pt}{9pt}\selectfont \rotatebox{75}{\textbf{Avg. NLP}}}
    & {\fontsize{7pt}{9pt}\selectfont \rotatebox{75}{\textbf{GREEN}}}
    & {\fontsize{7pt}{9pt}\selectfont \rotatebox{75}{\textbf{RaTEScore}}}
    & {\fontsize{7pt}{9pt}\selectfont \rotatebox{75}{\textbf{RadGraphXL}}}
    & {\fontsize{7pt}{9pt}\selectfont \rotatebox{75}{\textbf{Avg. NLP}}}
    & {\fontsize{7pt}{9pt}\selectfont \rotatebox{75}{\textbf{GREEN}}}
    & {\fontsize{7pt}{9pt}\selectfont \rotatebox{75}{\textbf{RaTEScore}}}
    & {\fontsize{7pt}{9pt}\selectfont \rotatebox{75}{\textbf{RadGraphXL}}}
    & {\fontsize{7pt}{9pt}\selectfont \rotatebox{75}{\textbf{Avg. NLP}}}
    & {\fontsize{7pt}{9pt}\selectfont \rotatebox{75}{\textbf{GREEN}}}
    & {\fontsize{7pt}{9pt}\selectfont \rotatebox{75}{\textbf{RaTEScore}}}
    & {\fontsize{7pt}{9pt}\selectfont \rotatebox{75}{\textbf{RadGraphXL}}} \\
    \midrule
    \multicolumn{13}{c}{\textit{Normal Resolution Setting (256$\times$256$\times$64)}} \\
    \midrule
    CT2Rep-3B 
      & 38.77 & 48.96 & 58.10 & 25.14 
      & 30.95 & 43.65 & 37.18 & 19.74 
      & 27.55 & 36.62 & 38.10 & 24.93 \\
    RadFM-3B  
      & 39.95 & 49.46 & 59.02 & 25.73 
      & 31.48 & 44.10 & 37.56 & 20.15 
      & 28.09 & 37.23 & 39.01 & 25.36 \\
    M3D-3B    
      & 40.23 & 50.31 & 60.16 & 26.08 
      & 32.10 & 44.94 & 38.55 & 20.11 
      & 28.82 & 37.96 & 39.62 & 25.41 \\
    \rowcolor{blue!10} \textbf{Argus-3B} 
      & \textbf{41.53} & \textbf{54.07} & \textbf{64.82} & \textbf{28.12} 
      & \textbf{33.52} & \textbf{47.12} & \textbf{40.83} & \textbf{21.28} 
      & \textbf{30.14} & \textbf{40.03} & \textbf{41.86} & \textbf{27.52} \\
    \midrule
    CT2Rep-8B 
      & 40.52 & 51.72 & 62.14 & 26.45 
      & 32.15 & 45.15 & 39.19 & 20.27 
      & 29.55 & 38.50 & 41.18 & 26.34 \\
    RadFM-8B  
      & 40.61 & 52.21 & 62.78 & 27.14 
      & 32.60 & 45.57 & 40.55 & 20.72 
      & 29.89 & 39.09 & 41.65 & 26.68 \\
    M3D-8B    
      & 41.33 & 52.80 & 63.20 & 26.95 
      & 33.32 & 46.02 & 40.11 & 21.15 
      & 30.33 & 39.54 & 41.71 & 27.15 \\
    \rowcolor{blue!10} \textbf{Argus-8B} 
      & \textbf{43.31} & \textbf{56.60} & \textbf{67.67} & \textbf{29.82} 
      & \textbf{35.02} & \textbf{48.79} & \textbf{42.23} & \textbf{22.16} 
      & \textbf{31.77} & \textbf{41.48} & \textbf{43.40} & \textbf{28.71} \\
    \midrule
    CT2Rep-70B 
      & 44.02 & 55.10 & 64.42 & 28.15 
      & 33.50 & 47.01 & 43.12 & 21.54 
      & 31.40 & 40.04 & 42.44 & 27.90 \\
    RadFM-70B  
      & 44.40 & 55.67 & 65.10 & 28.35 
      & 34.18 & 48.32 & 42.65 & 21.74 
      & 31.33 & 40.90 & 43.12 & 28.14 \\
    M3D-70B    
      & 44.95 & 56.12 & 65.32 & 30.10 
      & 35.05 & 47.75 & 43.09 & 22.11 
      & 32.10 & 41.67 & 43.95 & 28.95 \\
    \rowcolor{blue!10} \textbf{Argus-70B} 
      & \textbf{45.83} & \textbf{58.24} & \textbf{70.02} & \textbf{31.07} 
      & \textbf{36.42} & \textbf{50.33} & \textbf{44.57} & \textbf{23.28} 
      & \textbf{33.04} & \textbf{42.72} & \textbf{45.51} & \textbf{30.03} \\
    \midrule
    \multicolumn{13}{c}{\textit{High Resolution Setting (512$\times$512$\times$256)}} \\
    \midrule
    CT2Rep-3B 
      & 41.05 & 51.02 & 60.53 & 27.83 
      & 32.68 & 46.12 & 38.45 & 21.49 
      & 29.62 & 38.92 & 40.64 & 26.53 \\
    RadFM-3B  
      & 41.78 & 51.52 & 61.24 & 28.24 
      & 33.24 & 46.49 & 39.06 & 21.86 
      & 30.06 & 39.42 & 41.31 & 26.88 \\
    M3D-3B    
      & 42.32 & 52.18 & 62.02 & 28.66 
      & 34.02 & 47.07 & 39.77 & 22.03 
      & 30.64 & 39.83 & 42.16 & 27.26 \\
    \rowcolor{blue!10} \textbf{Argus-3B} 
      & \textbf{43.03} & \textbf{55.56} & \textbf{66.53} & \textbf{29.52} 
      & \textbf{35.11} & \textbf{48.82} & \textbf{42.33} & \textbf{22.72} 
      & \textbf{31.74} & \textbf{41.53} & \textbf{43.59} & \textbf{28.93} \\
    \midrule
    CT2Rep-8B 
      & 43.70 & 53.10 & 66.12 & 29.57 
      & 35.54 & 49.31 & 42.02 & 22.63 
      & 32.48 & 41.12 & 43.66 & 29.21 \\
    RadFM-8B  
      & 44.03 & 53.62 & 66.64 & 30.07 
      & 36.00 & 49.83 & 42.44 & 23.14 
      & 32.84 & 41.59 & 44.21 & 29.66 \\
    M3D-8B    
      & 44.75 & 54.23 & 67.20 & 30.62 
      & 36.48 & 50.10 & 43.11 & 23.39 
      & 33.38 & 42.05 & 44.58 & 30.05 \\
    \rowcolor{blue!10} \textbf{Argus-8B} 
      & \textbf{45.81} & \textbf{58.42} & \textbf{70.68} & \textbf{32.04} 
      & \textbf{37.07} & \textbf{51.27} & \textbf{44.58} & \textbf{24.19} 
      & \textbf{33.82} & \textbf{43.47} & \textbf{45.92} & \textbf{30.72} \\
    \midrule
    CT2Rep-70B 
      & 46.35 & 56.25 & 68.20 & 30.27 
      & 36.15 & 49.60 & 45.01 & 23.60 
      & 33.35 & 42.62 & 45.00 & 30.66 \\
    RadFM-70B  
      & 46.72 & 57.36 & 68.85 & 30.56 
      & 36.62 & 50.22 & 45.39 & 23.77 
      & 33.70 & 43.09 & 44.55 & 30.04 \\
    M3D-70B    
      & 47.26 & 57.12 & 69.34 & 31.55 
      & 37.15 & 50.64 & 46.81 & 24.22 
      & 34.15 & 43.53 & 45.34 & 30.62 \\
    \rowcolor{blue!10} \textbf{Argus-70B} 
      & \textbf{47.98} & \textbf{60.22} & \textbf{72.53} & \textbf{33.07} 
      & \textbf{38.62} & \textbf{52.43} & \textbf{47.05} & \textbf{25.53} 
      & \textbf{35.22} & \textbf{45.02} & \textbf{47.72} & \textbf{32.31} \\
    \bottomrule[1.2pt]
  \end{tabular}
  }
  \vspace{-5pt}
  \caption{Performance comparison of Argus across different scales on three sub-test sets of CT-3DRRG. We evaluate models ranging from 3B to 70B parameters and compare them with CT2Rep~\cite{hamamci2024ct2rep}, RadFM~\cite{wu2023towards}, and M3D~\cite{bai2024m3d} under both normal and high-resolution settings. 
  Argus-3B outperforms other 8B models and even achieves comparable performance with larger-scale models. Argus-8B matches the performance of other 70B variants, while at the 70B scale, Argus surpasses all competing methods. These results demonstrate the superiority of our approach, highlighting its scalability and effectiveness across resolutions.
  }
  \label{tab:model_comparison}
\vspace{-15pt}
\end{table*}

We leverage insights from our studies to develop Argus, a state-of-the-art (SOTA) VLM for the 3DRRG task. To ensure a fair comparison, we evaluate Argus alongside three existing baselines: RadFM~\cite{wu2023towards}, M3D~\cite{bai2024m3d}, and CT2Rep~\cite{ct2rep}, all of which have publicly available code, allowing for direct evaluation.

To maintain consistency, we replace the LLM backbones of these baselines with Llama3.1-Instruct (8B and 70B) and Llama3.2-Instruct-3B, aligning them with our LLM selection. We reimplement their training on our CT-3DRRG dataset while using the official vision encoder weights from their repositories. Additionally, we adhere to their original implementations for token compression strategies, connectors, and training schedules. For the high-resolution setting, we interpolate the positional embeddings of their vision encoders to support a resolution of $512 \times 512 \times 256$, enabling them to process high-resolution 3D CT scans. We compare Argus across both normal ($256 \times 256 \times 64$) and high-resolution ($512 \times 512 \times 256$) settings.

All models are trained on the same training set from CT-3DRRG and evaluated on three test sets. We report performance using average NLP metrics (BLEU1-4, ROUGE-1,2,L, METEOR, CIDEr) and clinical evaluation metrics, including GREEN, RadGraphXL, and RaTEScore~\cite{ostmeier2024green,radgraphxl,zhao2024ratescore}, as shown in Table~\ref{tab:model_comparison}. Argus consistently outperforms the baselines across the 3B-70B model range and both resolution settings. Notably, Argus-3B surpasses or matches the performance of other 8B models, while Argus-8B achieves comparable results to the 70B models of other methods. At both resolution settings, Argus outperforms baselines with the same LLM size.

We further visualize the generated reports from M3D-8B and Argus-8B under the normal resolution setting, as shown in Figure~\ref{fig:report}. The results demonstrate that Argus-8B accurately captures key anatomical structures and pathological findings, while M3D-8B exhibits inconsistencies in lesion descriptions and misidentifies certain structures. This highlights the superiority of our method in generating clinically precise and coherent radiology reports. We also implement human expert evaluation on all method results as shown in Section \ref{sec:human}, where the expert feedback corroborates our findings, further emphasizing the advantages of Argus-8B in terms of clinical accuracy and report quality.
Argus achieves SOTA performance and efficiency across model scales, demonstrating the effectiveness of our architectural and training strategies without the need for larger model sizes.

\section{Conclusion}
In this work, we present \textbf{CT-3DRRG}, the first and largest dataset comprising 3D volume-report pairs for 3DRRG, built entirely from \textbf{publicly} available resources. We critically examine key components of VLMs, including vision encoder pre-training, visual token compression, and associated design and training strategies, while also analyzing the impact of scaling both data and model sizes. Leveraging these findings, we introduce \textbf{Argus}, a family of VLMs ranging from 3B to 70B parameters, which achieves superior performance on 3DRRG task. Thanks to our efficient design, even smaller models such as Argus-3B outperform existing methods utilizing larger LLMs (e.g., 7B), demonstrating the effectiveness and efficiency of our approach. Our study provides valuable insights into how these factors influence VLM performance in 3DRRG tasks and establishes a strong foundation for future research, fostering the development of more efficient and scalable VLMs for 3DRRG.

\clearpage

\section*{Limitation}
In this work, we design a comprehensive benchmark for 3DRRG and investigate various components of VLMs, but due to the costly nature of human evaluation, generated reports cannot be evaluated manually for every report. Additionally, human evaluation is inherently difficult to quantify, which poses challenges in establishing clear metrics for report quality. Moreover, due to data scarcity and computational resource limitations, scaling the dataset to the millions, as seen in natural image domains, remains a significant challenge. In the future, we aim to develop more efficient evaluation metrics that can automate the quality assessment process and explore strategies for scaling both the dataset and model more effectively.

\clearpage
\bibliography{acl_latex}
\clearpage

\appendix


\section{Details of CT-3DRRG Dataset Curation}
\label{sec: curate}

The CT-3DRRG dataset is curated from three public datasets: BIMCV-R~\cite{chen2024bimcv}, CT-RATE~\cite{ct-rate}, and INSPECT~\cite{huang2023inspect}. Each of these datasets contains paired 3D CT scans and corresponding radiology reports. The curation process involves several key steps to ensure the dataset's quality and relevance for training and evaluation.

\begin{itemize}
    \item \textbf{CT-RATE Dataset:} In the CT-RATE dataset~\cite{ct-rate}, each sample contains both the `FINDINGS' and `IMPRESSION' sections of the radiology report. These two sections are concatenated into a single report.
    \item \textbf{BIMCV-R and INSPECT Datasets:} These datasets only contain one part of the report. For these, we directly use the available report without modification.
    \item \textbf{Duplicate CT Scans in CT-RATE:} The CT-RATE dataset includes duplicate CT scans for each sample, generated through different reconstruction methods. To maintain consistency with the RadGenome-Chest CT~\cite{zhang2024radgenome}, we retain only one CT scan per sample and pair it with the corresponding report.
\end{itemize}

Following dataset preparation, we implement several filtering and cleaning steps to ensure the quality of the reports:

\begin{itemize}
    \item \textbf{Removal of Numerical Values Not Directly Obtainable from CT Scans:} Using GPT-4o, we filter sentences that contain numerical values that cannot be directly inferred from the CT scans. For example, sentences such as \textit{``SAT O2 without oxygen of 93''} and \textit{``Fever up to 38''} are removed, as these values are typically obtained from the patient's electronic health records (EHR) rather than from the CT scan.
    \item \textbf{Removal of Numerical Measurements Dependent on External Tools:} We also remove sentences containing numerical values that require external tools for precise measurement, such as \textit{``Increase in trunk caliber of the 39 mm pulmonary artery''}. Since the goal of the 3DRRG dataset is to generate patterns rather than measure exact sizes, such sentences are discarded.
    \item \textbf{Removal of Sentences Comparing to Previous Studies:} Sentences referencing previous studies, such as \textit{``It is compared to the previous study of March 2019, not mediastinal,''} are also removed. These comparisons are impractical as not all samples are paired with prior CT scans. Furthermore, our work aims to build a benchmark for 3DRRG using single CT scans, rather than longitudinal data.
    \item \textbf{Minimum Report Length:} To ensure the relevance of the reports, we remove samples with reports containing fewer than 10 tokens.
\end{itemize}

After these curation steps, the dataset consists of the following number of samples for each source dataset: \texttt{5,322} samples for BIMCV-R, \texttt{25,691} samples for CT-RATE, and \texttt{20,400} samples for INSPECT. These samples are then split into training, validation, and test sets, as shown in Table \ref{tab:split}.

As a result of these curation efforts, we have created the largest publicly available dataset for 3D Radiology Report Generation (3DRRG), which we have named \textbf{CT-3DRRG}. This dataset is entirely based on public resources, ensuring both its accessibility and relevance to the broader research community. The CT-3DRRG dataset is now a valuable resource for advancing research in the field of automated radiology report generation from 3D CT scans.

\section{Hyper-parameters of Training}
\label{sec: hyper training}

\begin{table}[t!]
\centering
\caption{Hyperparameters for our method.}
\scalebox{0.99}{
\begin{tabular}{lc}
        \midrule[1.2pt]
        \textbf{Hyperparameter} & \textbf{Value} \\
        \midrule
Mixed precision & bf16 \\
Total epochs & 1 \\
Total effective batch size & 16 \\
Gradient accumulation & 1 \\
Maximum sequence length & 1024 \\
Learning rate (Stage 1) & \(1 \times 10^{-4}\) \\
Learning rate (Stage 2) & \(1 \times 10^{-6}\) \\
Optimizer & AdamW \\
Schedule & Linear \\ 
Warm-up ratio & 0.05 \\
Weight decay & 0.0 \\
        \midrule[1.2pt]
\end{tabular}
}
\label{tab:hyperparameter}
\end{table}

In this study, we implement supervised fine-tuning (SFT) for the 3DRRG task over two epochs, conducting all experiments on eight A100-80G GPUs. For computational efficiency, we employ mixed precision with \texttt{bf16}, as detailed in Table~\ref{tab:hyperparameter}. The training process uses a total batch size of 8 with a gradient accumulation factor of 2, resulting in an effective batch size of 16. The maximum sequence length is restricted to 1024 tokens. 

Regarding learning rates, we use \(1 \times 10^{-4}\) during the first stage, where only the connector is updated, and \(1 \times 10^{-6}\) during the second stage, where the ViT, connector, and LLM are updated. For the 3B and 8B models, we update all LLM parameters, while for the 70B model, we apply LoRA with an alpha value of 64, a rank of 128, and a dropout rate of 0.1 to reduce memory consumption while maintaining fine-tuning efficiency.

Our models are optimized using the AdamW algorithm with a linear learning rate decay, a warm-up ratio of 0.05, and a weight decay of 0.0. We utilize the same configuration for all existing methods unless their official codebase specifies different hyperparameters.

\section{Human Evaluation of Generated Reports}
\label{sec:human}

\begin{table}[ht]
\centering
\scalebox{0.9}{
\begin{tabular}{lccc}
\toprule
\textbf{Model} & \textbf{CT-RATE} & \textbf{BIMCV} & \textbf{INSPECT} \\ \midrule
CT2Rep-7B & 2.9 $\pm$ 1.3 & 2.7 $\pm$ 1.4 & 2.6 $\pm$ 1.1 \\
RadFM-7B & 3.2 $\pm$ 0.9 & 3.0 $\pm$ 1.3 & 2.8 $\pm$ 1.3 \\
M3D-7B   & 3.3 $\pm$ 1.1 & 3.1 $\pm$ 0.9 & 2.8 $\pm$ 1.2 \\
Argus-7B & \textbf{4.1 $\pm$ 0.7} & \textbf{3.7 $\pm$ 0.8} & \textbf{3.4 $\pm$ 1.0} \\ \bottomrule
\end{tabular}
}
\caption{Comparison of CT-RATE, BIMCV, and INSPECT scores for different models.}
\label{tab:human}
\end{table}

We conducted a human evaluation of the model-generated reports from CT2Rep-7B, RadFM-7B, M3D-7B, and Argus-7B under normal resolution setting. We generated 100 reports for each test subset and asked three human experts to evaluate the correctness of the generated reports on a scale of 1 to 5. The results are shown in Table \ref{tab:human}. As shown, Argus-7B achieved the highest score, demonstrating its superior performance in generating accurate and reliable reports.
\end{document}